\documentclass[10pt,twocolumn,letterpaper]{article}

 \usepackage{cvpr}              
\usepackage{caption}
\usepackage{graphicx}
\usepackage{float}
\usepackage{subcaption}
\usepackage{paralist}
\usepackage{multirow}
\usepackage{booktabs}
\usepackage{pifont}
\usepackage{bbding}
\makeatletter
\def\hlinew#1{%
	\noalign{\ifnum0=`}\fi\hrule \@height #1 \futurelet
	\reserved@a\@xhline}
\makeatother%

\usepackage[dvipsnames]{xcolor}

\definecolor{cvprblue}{rgb}{0.21,0.49,0.74}
\usepackage[pagebackref,breaklinks,colorlinks,citecolor=cvprblue]{hyperref}

\def\paperID{1950} 
\def\confName{CVPR}
\def\confYear{2024}

\title{Bidirectional Multi-Scale Implicit Neural Representations for Image Deraining}

\author{Xiang Chen \quad Jinshan Pan\thanks{Corresponding author.} \quad Jiangxin Dong\\
Nanjing University of Science and Technology \\
	\\
}

\begin{document}
\maketitle
\begin{abstract}
How to effectively explore multi-scale representations of rain streaks is important for image deraining.
In contrast to existing Transformer-based methods that depend mostly on single-scale rain appearance, we develop an end-to-end multi-scale Transformer that leverages the potentially useful features in various scales to facilitate high-quality image reconstruction.
To better explore the common degradation representations from spatially-varying rain streaks, we incorporate intra-scale implicit neural representations based on pixel coordinates with the degraded inputs in a closed-loop design, enabling the learned features to facilitate rain removal and improve the robustness of the model in complex scenarios.
To ensure richer collaborative representation from different scales, we embed a simple yet effective inter-scale bidirectional feedback operation into our multi-scale Transformer by performing coarse-to-fine and fine-to-coarse information communication.
Extensive experiments demonstrate that our approach, named as NeRD-Rain, performs favorably against the state-of-the-art ones on both synthetic and real-world benchmark datasets.
The source code and trained models are available at \url{https://github.com/cschenxiang/NeRD-Rain}.
\end{abstract}
\vspace{-4mm}

\section{Introduction}
\label{sec:intro}
\vspace{-2mm}
Recent years have witnessed significant progress in image deraining due to the development of numerous deep convolutional neural networks (CNNs)~\cite{fu2017removing,zhang2018density,wang2019spatial,jiang2020multi,yi2021structure}.
However, as the basic operation in CNNs, the
convolution is spatially invariant and has limited
receptive fields, which cannot effectively model the spatially-variant property and non-local structures of clear images~\cite{zamir2022restormer,vaswani2017attention}. Moreover, simply increasing the network depth to obtain larger receptive fields does not always lead to better performance.

To alleviate this problem, several recent approaches utilize Transformers to solve single image deraining \cite{guo2023sky,chen2023learning,zamir2022restormer,xiao2022image,wang2022uformer,chen2021pre}, since Transformers can model the non-local information for better image restoration.
Although these approaches achieve better performance than most of the CNN-based ones, they mostly explore feature representations at a fixed image scale (\ie, a single-input single-output architecture), while ignoring potentially useful information from other scales.
As the rain effect decreases significantly at coarser image scales, exploring the multi-scale representation would facilitate the rain removal.

\begin{figure}[!t]\footnotesize
	\centering
	\begin{tabular}{ccc}
 \hspace{-2mm}
		\includegraphics[width=0.325\linewidth]{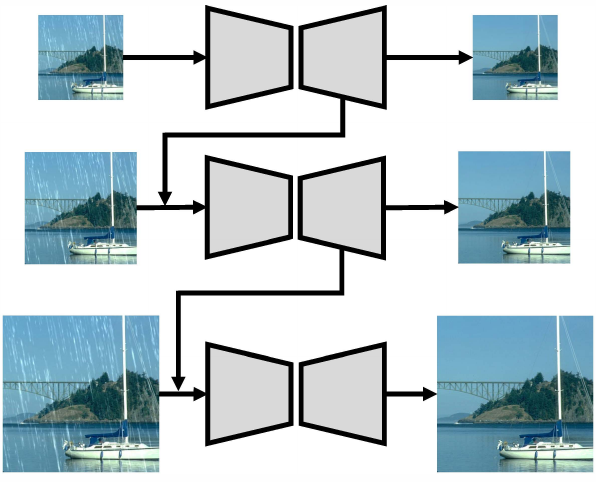} &\hspace{-4mm}
		\includegraphics[width=0.285\linewidth]{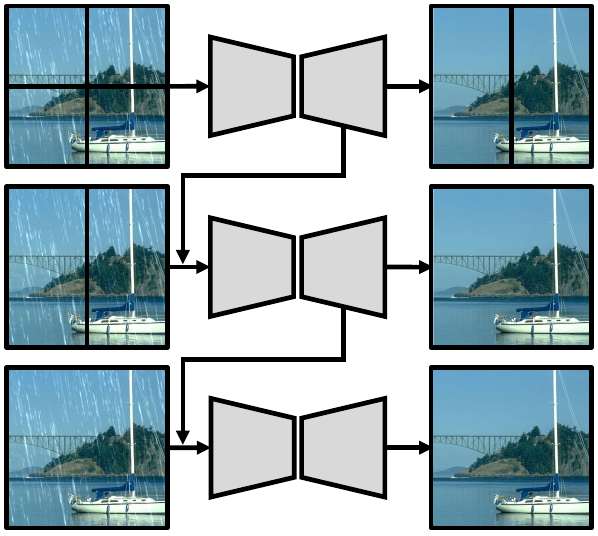} &\hspace{-4mm}
		\includegraphics[width=0.35\linewidth]{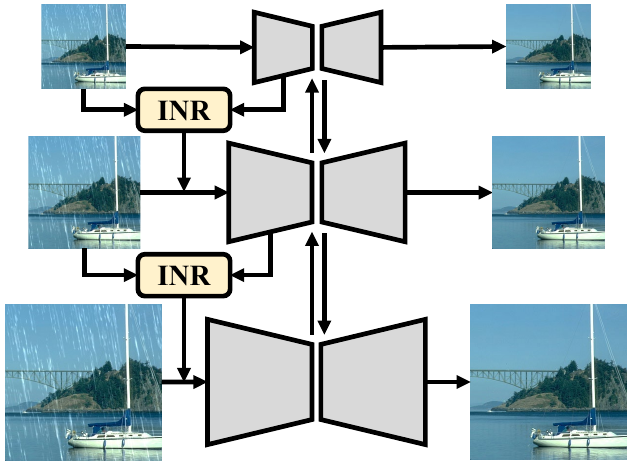}  \\
		(a) Coarse-to-fine &\hspace{-4mm}  (b) Multi-patch & \hspace{-4mm} (c)  Ours \\
	\end{tabular}
	\vspace{-3mm}
	\caption{Illustration of the proposed approach and the currently existing multi-scale solutions. (a) coarse-to-fine scheme~\cite{zhang2018density,jiang2020multi}; (b) multi-patch scheme~\cite{zamir2021multi}; (c) our method. Compared to previous approaches, the method one integrates implicit neural representations (INR) into our bidirectional multi-scale model to form a closed-loop framework, which allows for better exploration of multi-scale information and modeling of complex rain streaks.}
	\vspace{-5mm}
	\label{fig1}
\end{figure}

To this end, several approaches introduce the coarse-to-fine mechanism~\cite{tao2018scale, cho2021rethinking} or multi-patch strategy~\cite{zamir2021multi} into deep neural networks to exploit multi-scale rain features.
As shown in Figure~\ref{fig1}, the decoder's feature or derained image is initially estimated at a coarse scale and then used as the input at a finer scale for guidance.
Although impressive performance has been achieved, these methods are less effective when handling complex and random rain streaks because these rain streaks cannot be removed by downsampling operations and inaccurate estimation of a coarser scale will result in suboptimal restoration performance at finer scales.
Despite the spatially-varying rain streaks exhibit a variety of scale properties (\eg, size, shape, length, and density), the degraded rainy images tend to share some similar visual degradation characteristics (\ie, common rain degradation representation)~\cite{wang2017hierarchical,xiao2022image,wang2020rethinking}.
However, existing methods do not effectively model the common degradation as they usually rely on traditional representation forms that are sensitive to the input variation rather than capturing underlying implicit functions, which limits their performance on complex scenarios.
Thus, it is of great interest to learn the underlying correlations among features to encode rain appearance information from spatially-varying rain streaks.

Furthermore, we note that most existing multi-scale architectures~\cite{tao2018scale,jiang2020multi,kim2022mssnet,li2022deep} utilize the features from coarser scales to facilitate the feature estimation at finer scales.
However, if the features are not estimated correctly at coarser scales, the errors would affect the feature estimation at subsequent scales.
Therefore, it is necessary to introduce a feedback mechanism to solve this problem.

In this paper, we develop an effective bidirectional multi-scale Transformer with implicit neural representations to better explore multi-scale information and model complex rain streaks.
Considering that the rain effect varies at different image scales, we construct multiple unequal Transformer branches each for learning the scale-specific features for image deraining.
Motivated by the recent success of implicit neural representations (INRs) that are able to encode an image as a continuous function, we further incorporate the INR between adjacent branches to learn common rain degradation representations from diverse degraded inputs so that the learned features are robust to complex and random rain streaks.
To facilitate representing rain appearance at various scales, we employ two distinct coordinate-based multi-layer perceptrons (MLPs) (\ie, one coarse and one fine feature grid) in INR for adaptively fitting complex rain characteristics.
Furthermore, to improve the modeling capacity of the INR, we propose an intra-scale shared encoder to form a closed-loop framework.
Note that the above mentioned two types of representation (\ie, scale-specific and common rain ones) are able to complement each other.

To better build the interactions among features of different scales in a collaborative manner, we introduce a simple yet effective inter-scale bidirectional feedback mechanism into our proposed multi-scale Transformer.
This enables the network to flexibly exchange information in both coarse-to-fine and fine-to-coarse flows, resulting in better robustness to variations in image content, such as changes in scale.
Finally, we formulate the intra-scale INR branch and the inter-scale bidirectional branch into an end-to-end trainable image deraining network, named as NeRD-Rain.
Experimental results demonstrate that our approach achieves favorable performance against state-of-the-art ones on the
benchmark datasets, especially on real-world benchmarks.

The main contributions are summarized as follows:
\begin{compactitem}
\item We design an effective multi-scale Transformer to generate high-quality deraining results with exploitation and utlization of multi-scale rain information.
	
\item We introduce implicit neural representations to better learn common rain degradation features and show that it can help facilitate rain removal and enhance the robustness of the deraining model in complex scenes.
	
\item We integrate a simple yet effective bidirectional feedback propagation operation into our multi-scale Transformer for better feature interaction across scales.
	
\item Experimental results on both synthetic and real-world benchmarks demonstrate that our approach performs favorable performance against state-of-the-art ones.
\end{compactitem}

\section{Related Work}
\label{sec:related}
\vspace{-2mm}
{\flushleft\textbf{Single image deraining}.}
Since the image deraining problem is ill-posed, conventional algorithms typically seek to impose some handcrafted priors~\cite{kang2011automatic,luo2015removing,li2016rain,zhu2017joint,wang2017hierarchical,gu2017joint} on the clear images and rain components to make this problem well-posed.
However, these methods fail to remove rain on complex real scenes when the assumptions do not hold.
Afterwards, deep learning-based approaches~\cite{ba2022not,chen2023towards,chen2022unpaired} have outperformed early traditional algorithms and demonstrated decent restoration performance.
As the development of this field, a wide array of network architectures and designs have been effectively explored to significantly boost the capacity of end-to-end model learning, \eg, multi-scale~\cite{jiang2020multi,li2022deep}, multi-stage~\cite{li2018recurrent,zamir2021multi}, or multi-branch~\cite{zhang2018density,wang2020rethinking} strategy.

Recently, researchers have made efforts to replace CNNs with Transformers as the fundamental structure for vision tasks~\cite{dosovitskiy2020image,chen2023towards}.
Driven by the great success of vision Transformer in modeling the non-local information, Transformer-based frameworks~\cite{zamir2022restormer,chen2021pre} have emerged for better rain removal.
For example, Xiao \emph{et al.}~\cite{xiao2022image} proposed an image deraining Transformer using spatial-based and
window-based self-attention modules.
Chen \emph{et al.}~\cite{chen2023learning} put forward a sparse Transformer to retain the most useful self-attention values for image reconstruction.
Unfortunately, these approaches depend mostly on single-scale rain appearance, which limits their ability to fully explore multi-scale rain information.
In this work, we investigate multi-scale representations in Transformer backbone for better boosting image deraining.

\vspace{-2.5mm}

{\flushleft\textbf{Multi-scale representations}.}
The presence of rain streaks exhibits a noticeable degree of self-similarity, whether it is within the same scale or across different scales. This inherent property allows for the utilization of correlated features across scales in order to better represent rain information~\cite{jiang2020multi,li2022deep,chen2023towards,zhang2018density}.
In the context of CNNs, multi-scale representations have been explored for improving the image restoration performance, such as image pyramid~\cite{fu2019lightweight}, feature pyramid~\cite{mei2023pyramid}, coarse-to-fine mechanism~\cite{tao2018scale, cho2021rethinking}, and multi-patch scheme~\cite{zamir2021multi}.
Recently, some studies~\cite{fan2021multiscale} have investigated enforcing multi-scale design strategies to vision Transformers (ViT).
For example, Chen \emph{et al.}~\cite{chen2021crossvit} formulated a cross-attention multi-scale Transformer for image classification.
Lin \emph{et al.}~\cite{lin2023scale} developed a scale-aware modulation Transformer.
%
In this work, we formulate a bidirectional multi-scale Transformer to solve image deraining.

\begin{figure*}[t]
	\centering
	\includegraphics[width=1.0\textwidth]{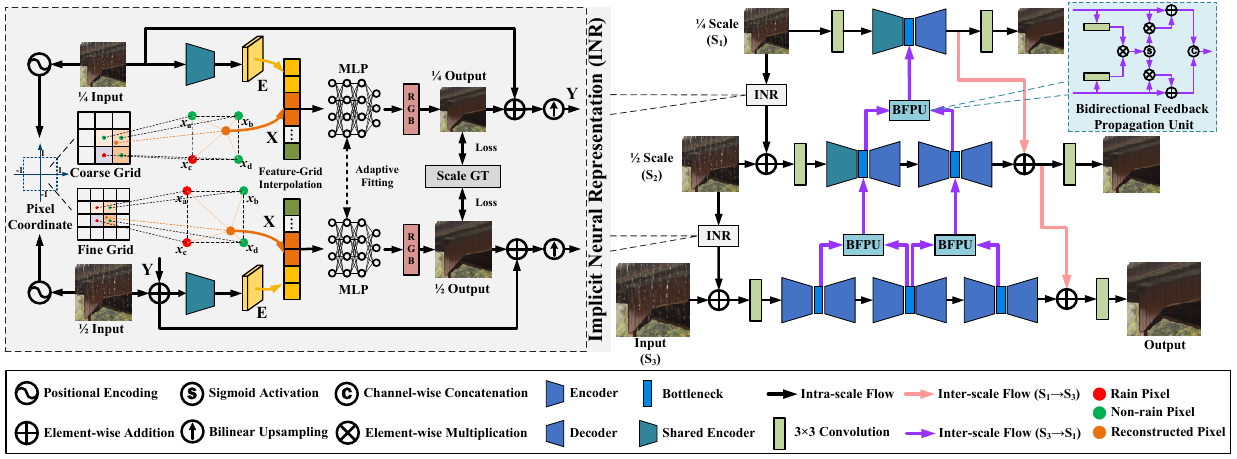}
	\vspace{-6mm}
	\caption{Overall architecture of the proposed bidirectional multi-scale Transformer with implicit neural representations (NeRD-Rain), which consists of intra-scale flows (\ie, INR branch and unequal Transformer branch) and inter-scale flows (\ie, coarse-to-fine and fine-to-coarse bidirectional branches). The proposed INR branch consists of two coordinated-based MLPs with coarse and fine feature grids. We construct an intra-scale shared encoder in the Transformer branch and INR branch, where two types of representation (\ie, scale-specific and common rain ones) are able to complement each other. We formulate all the branches to form a closed-loop network architecture.}
	\label{fig2}
	\vspace{-5mm}
\end{figure*}

\vspace{-2.5mm}

{\flushleft\textbf{Implicit neural representation}.}
Implicit neural representation (INR) has emerged as a new compelling technique to represent continuous domain signals via coordinate-based multi-layer perceptrons (MLPs).
Unlike explicit representations, which define the signal values at each point explicitly, INR encodes the signal by learning a mapping from coordinates to signal values \cite{saragadam2022miner}.
During the early period, it has been widely applied in various 3D vision tasks, \eg, shape modeling \cite{chen2019learning}, structure rendering \cite{barron2021mip}, and scene reconstruction \cite{jiang2020local}.
As a well-known approach, neural radiance fields (NeRF) \cite{mildenhall2021nerf} employs neural networks to represent complex 3D scenes by modeling the volumetric density and color at each point in space.
Recent studies attempt to explore the potential of INRs for 2D images, \eg, image compression \cite{strumpler2022implicit}, image reconstruction \cite{xu2022signal,nam2022neural}, and arbitrary-scale image super-resolution \cite{chen2021learning,chen2023cascaded}.
More recently, Yang \emph{et al.} \cite{yang2023implicit} utilized the controllable fitting capability of INR for low-light image enhancement problem.
Quan \emph{et al.} \cite{quan2023single} proposed an INR-based inverse kernel prediction network to solve image defocus deblurring.
Our work is inspired by this rapidly growing field and further demonstrates how to use implicit representation to better facilitate rain removal.

\vspace{-2mm}

\section{Proposed Method}
\label{sec:method}
\vspace{-2mm}
To better explore multi-scale information and model complex rain streaks, we elaborately develop an effective bidirectional multi-scale Transformer with implicit neural representations (called NeRD-Rain), comprising an intra-scale INR branch and an inter-scale bidirectional branch.
The former learns the underlying degradation representations from diverse rainy images, while the latter enables richer collaborative representations across different scales.
Figure~\ref{fig2} summarizes the architecture of NeRD-Rain.

\vspace{-1.5mm}

\subsection{Intra-scale INR branch}
\label{sec:3.1}
\vspace{-1mm}
Given a rainy image $I_{rain} \in \mathbb{R}^{H \times W \times 3}$, where $H \times W$ represents the spatial resolution of the input image, our method first uses bilinear interpolation to downsample the input image into multi-scale versions (\ie, 1/2 and 1/4).
From the coarsest to the finest image scales, we designate the rescaled image at each scale as $S_{1}$, $S_{2}$, and $S_{3}$, respectively.
%
%
Different from previous multi-scale methods~\cite{tao2018scale,jiang2020multi,li2022deep} that assign equal importance to the subnetworks of various scales, our approach incorporates the networks at finer scales with deeper architectures to handle spatially-varying rain streaks.
At each scale, we propose unequal Transformer branches to perform deep feature extraction and generate a set of scale-specific outputs.
Specifically, the NeRD-Rain at each scale (from $S_{1}$ to $S_{3}$) is equipped with one, two, and three UNets, respectively.
Each UNet consists of a sequence of Transformer blocks~\cite{zamir2022restormer}.
Here, these UNets share the same network architecture but have independent weights~\cite{kim2022mssnet}.

In order to capture common rain degradation features, we further integrate an intra-scale INR branch into our multi-scale Transformer, which trains a multi-layer perceptron (MLP) by learning the following mapping function:
\vspace{-2mm}
\begin{equation}
\vspace{-2mm}
f_\theta: \mathbb{R}^2 \rightarrow \mathbb{R}^3,
\end{equation}
where the input dimensions correspond to the $(x, y)$ spatial coordinates of each pixel, while the output dimensions correspond to the $(R, G, B)$ color channels of the pixels.

Specifically, we insert INRs between the inputs of adjacent Transformer branches to synchronously achieve rain reconstruction.
Firstly, the input image $I_{rain}$ is transformed to a feature map $\mathbf{E} \in \mathbb{R}^{H \times W \times C}$ with a spatial resolution of $H \times W$ pixels and $C$ channels.
Different from the INR in~\cite{yang2023implicit} which utilizes a separate encoder, we construct a shared encoder that interacts with the Transformer branch to form a compact closed-loop architecture, where these two types of representation (\ie, scale-specific and common rain representation) can be utilized complementary to each other.
In addition, the position of each pixel from spatially-varying rain streaks is recorded in a relative coordinate set $\mathbf{X} \in \mathbb{R}^{H \times W \times 2}$, where the value `2' represents horizontal and vertical coordinates.
As suggested in~\cite{mildenhall2021nerf,lee2022local}, we also adopt periodic spatial encoding to project the pixel coordinates $\mathbf{X}$ into a higher dimensional space $\mathbb{R}^{2 L}$ for better recovering high-frequency details.
The encoding procedure is formulated as:
\vspace{-1mm}
\begin{equation}
\vspace{-2mm}
\begin{split}
&\mathbf{X}^{\prime} = \gamma(\mathbf{X}),
\\
&\gamma(\mathbf{x})=\left[\sin (\mathbf{x}), \cos (\mathbf{x}), \ldots, \sin \left(2^{L-1} \mathbf{x}\right), \cos \left(2^{L-1} \mathbf{x}\right)\right],
\end{split}
\end{equation}
where $\gamma(\cdot)$ represents a spatial encoding function. $\mathbf{x} \in \mathbb{R}^{2L}$ is the coordinate value of $\mathbf{X}$, and it is normalized to lie within the range of $[-1,1]$. $L$ is a hyperparameter for determining dimension values. We set $L=4$ in our experiments.

Afterwards, a decoder is employed to predict RGB values of the output image by combining both $\mathbf{E}$ and $\mathbf{X}^{\prime}$. Here, our decoder consists of three-layer MLPs, with each layer having 256 hidden dimensions.
Note that fitting an INR to reconstruct an image requires finding a set of parameters for the MLP $f_\theta$ of a small size \cite{rivas2023ice}.
As a result, diverse types of rain streaks yield different sets of parameters, and it in turn means the MLP is adaptive to the common characteristics of all the degraded images.
Similar to~\cite{chen2021learning,yang2023implicit}, we calculate a weighted average of the predictions from the surrounding grids to obtain the RGB value ($\mathbf{s} \in \mathbb{R}^3$) of the final reconstructed image, which can be viewed as an implicit neural interpolation process \cite{tang2021joint,chen2023cascaded}.
This process is expressed as:
\vspace{-2mm}
\begin{equation}
\vspace{-3mm}
\begin{split}
&\mathbf{z}=\mathbb{E}(I_{rain}),
\\	
&\mathbf{s}=\sum_{j \in \mathcal{J}} w_j f_\theta\left(\mathbf{z}_j, \mathbf{x}\right),	
\end{split}
\end{equation}	
where $\mathbf{z}$ is a feature vector; $\mathbb{E}$ represents a shared encoder; $\mathcal{J} \in \mathbb{Z}^4$ is a set indices for four nearset (Euclidean distance) latent codes $j$ around $\mathbf{x}$; $w_j$ denotes the  local ensemble weight~\cite{lee2022local}, satisfying $\sum_j w_j=1$.

In experiments, we further find that this process can naturally facilitate rain removal without requiring any additional operations.
Likewise, some studies~\cite{kim2022zero,chen2021nerv} also point out the low-pass filtering characteristics in INR.
Due to the strong reflections caused by rain effect, pixels affected by rain tend to exhibit high intensity values, \ie, white rain streaks~\cite{wang2017hierarchical}.
Therefore, we attribute the deraining ability of INR to a basic fact that the intensity values of rain-affected pixels tend to surpass those of their neighboring non-rain pixels~\cite{wang2017hierarchical}.

Instead of representing the image using INR at a fixed scale~\cite{chen2023cascaded}, we present a cascaded scale image representation for INR.
Inspired by \cite{girish2023shacira}, our network trains two distinct MLPs, \ie, one coarse and one fine feature grid.
Through this sequential coarse-to-fine training, INR achieves more effective information transmission, naturally sharing information across scales.
With all the above-mentioned designs, our INR branch can better learn common rain degradation features so that the learned features are robust to complex and random rain streaks.
These designs we consider yield performance improvements as we shall see in Section \ref{sec:analysis}.


\subsection{Inter-scale bidirectional branch}
\vspace{-1mm}
Although the intra-scale INR branch performs feature estimation from coarse to fine scales, it would affect the feature estimation for the subsequent scale when the features from the coarser scales are not estimated correctly.
To overcome this problem, we introduce an inter-scale bidirectional branch into multi-scale Transformer, enabling both coarse-to-fine and fine-to-coarse feature propagation.
Specifically, unlike using the complex and time-consuming LSTM~\cite{quan2023single}, we formulate a simple yet effective bidirectional feedback propagation unit (BFPU) without adding much cost.
Each BFPU takes the bottleneck layer features ($F_{a}$ and $F_{b}$) of two UNets at the current fine scales as inputs.
The output of BFPU $F_{out}$ is delivered to the bottleneck layer of the UNet at the previous coarse scale.
In this way, the proposed BFPU can be formulated by:
\vspace{-1mm}
\begin{equation}
\vspace{-1mm}
\begin{split}
&F_{mid}=\mathcal{S}\left(\mathrm{Conv}_{3 \times 3}\left(F_{a}\right) \otimes \mathrm{Conv}_{3 \times 3}\left(F_{b}\right)\right),
\\
&F_{out}=[F_{a}+F_{mid} \otimes F_{a}, F_{b}+ F_{mid} \otimes F_{b}],
\end{split}
\end{equation}
where $\mathrm{Conv}_{3 \times 3}$ is a $3 \times 3$ convolution layer, $\mathcal{S}$ denotes the Sigmoid function, $\otimes$ is the element-wise multiplication, and $[\cdot]$ represents the channel-wise concatenation.

With this design, the inter-scale bidirectional branch offers three-fold advantages: (1) it can make use of the complementary information from the subsequent (finer) scales to help image restoration at the current (coarser) scale, (2) it can perform feature propagation flow earlier without waiting for the derained results from previous scales, (3) it can be robust to variations in image content, such as changes in scale.
We will show its effectiveness in Section \ref{sec:analysis}.

\begin{table*}[t]
	\centering
	\caption{Quantitative evaluations of the proposed approach against state-of-the-art methods on five commonly used benchmark datasets. Our NeRD-Rain achieves higher quantitative results, especially advances state-of-the-art by \textbf{1.04} dB on the real benchmark, SPA-Data.}
	\vspace{-2.5mm}
	\resizebox{1.0\textwidth}{!}{
		\begin{tabular}{cc|cc|cc|cc|cc|cc}
			\toprule
			\multicolumn{2}{c|}{Datasets}                                                   & \multicolumn{2}{c|}{Rain200L \cite{yang2017deep}}    & \multicolumn{2}{c|}{Rain200H \cite{yang2017deep}}    & \multicolumn{2}{c|}{DID-Data \cite{zhang2018density}}    & \multicolumn{2}{c|}{DDN-Data \cite{fu2017removing}}    & \multicolumn{2}{c}{SPA-Data \cite{wang2019spatial}} \\ \hline
			\multicolumn{2}{c|}{Metrics}                                                    & PSNR           & SSIM            & PSNR           & SSIM            & PSNR           & SSIM            & PSNR           & SSIM            & PSNR           & SSIM            \\ \hline
			\multicolumn{1}{c|}{\multirow{2}{*}{Prior-based methods}}       & DSC \cite{luo2015removing}          & 27.16          & 0.8663          & 14.73          & 0.3815          & 24.24          & 0.8279          & 27.31          & 0.8373          & 34.95          & 0.9416          \\
			\multicolumn{1}{c|}{}                                           & GMM \cite{li2016rain}          & 28.66          & 0.8652          & 14.50          & 0.4164          & 25.81          & 0.8344          & 27.55          & 0.8479          & 34.30          & 0.9428          \\ \hline
			\multicolumn{1}{c|}{\multirow{8}{*}{CNN-based methods}}         & DDN \cite{fu2017removing}          & 34.68          & 0.9671          & 26.05          & 0.8056          & 30.97          & 0.9116          & 30.00          & 0.9041          & 36.16          & 0.9457          \\
			\multicolumn{1}{c|}{}                                           & RESCAN \cite{li2018recurrent}       & 36.09          & 0.9697          & 26.75          & 0.8353          & 33.38          & 0.9417          & 31.94          & 0.9345          & 38.11           & 0.9707          \\
			\multicolumn{1}{c|}{}                                           & PReNet \cite{ren2019progressive}       & 37.80          & 0.9814          & 29.04          & 0.8991          & 33.17          & 0.9481          & 32.60          & 0.9459          & 40.16          & 0.9816          \\
			\multicolumn{1}{c|}{}                                           & MSPFN \cite{jiang2020multi}       & 38.58          & 0.9827          & 29.36          & 0.9034          & 33.72          & 0.9550          & 32.99          & 0.9333          & 43.43          & 0.9843          \\
			\multicolumn{1}{c|}{}                                           & RCDNet \cite{wang2020model}       & 39.17          & 0.9885          & 30.24          & 0.9048          & 34.08          & 0.9532          & 33.04          & 0.9472          & 43.36          & 0.9831          \\
			\multicolumn{1}{c|}{}                                           & MPRNet \cite{zamir2021multi}      & 39.47          & 0.9825          & 30.67          & 0.9110          & 33.99          & 0.9590          & 33.10          & 0.9347          & 43.64              & 0.9844               \\
			\multicolumn{1}{c|}{}                                           & DualGCN \cite{fu2021rain}       & 40.73          & 0.9886          & 31.15          & 0.9125          & 34.37          & 0.9620          & 33.01          & 0.9489          & 44.18          & 0.9902          \\
			\multicolumn{1}{c|}{}                                           & SPDNet \cite{yi2021structure}       & 40.50          & 0.9875          & 31.28          & 0.9207          & 34.57          & 0.9560          & 33.15          & 0.9457          & 43.20          & 0.9871          \\ \hline
			\multicolumn{1}{c|}{\multirow{6}{*}{Transformer-based methods}} & Uformer \cite{wang2022uformer}    & 40.20          & 0.9860          & 30.80          & 0.9105          & 35.02          & 0.9621          &  33.95         & 0.9545          & 46.13          & 0.9913          \\
			\multicolumn{1}{c|}{}                                           & Restormer \cite{zamir2022restormer}    & 40.99          & 0.9890          & 32.00          & 0.9329          & 35.29          & 0.9641          & 34.20          & 0.9571          & 47.98          & 0.9921          \\
			\multicolumn{1}{c|}{}                                           & IDT \cite{xiao2022image}          & 40.74          & 0.9884          & 32.10          & 0.9344          & 34.89          & 0.9623          & 33.84          & 0.9549          & 47.35          & 0.9930          \\
			\multicolumn{1}{c|}{}                                           & DRSformer \cite{chen2023learning} & 41.23 & 0.9894 & 32.17 & 0.9326 & 35.35 & 0.9646 & 34.35 & 0.9588 & 48.54 & 0.9924 \\
			\multicolumn{1}{c|}{}                                           & NeRD-Rain-S  & 41.30 & 0.9895 & 32.06 & 0.9315 & 35.36 & 0.9647 & 34.25 & 0.9578 & 48.90 & 0.9936 \\
			\multicolumn{1}{c|}{}                                           & NeRD-Rain  & \textbf{41.71} & \textbf{0.9903} & \textbf{32.40} & \textbf{0.9373} & \textbf{35.53} & \textbf{0.9659} & \textbf{34.45} & \textbf{0.9596} & \textbf{49.58} & \textbf{0.9940} \\ \bottomrule
		\end{tabular}
	}
	\vspace{-2.5mm}
	\label{table1}	
\end{table*}


\subsection{Loss function}
\vspace{-1mm}
In order to jointly learn UNet-based traditional representations and INR-based continuous representations in a multi-scale manner, our network is trained end-to-end with a hybrid loss function.
Following~\cite{cho2021rethinking,tu2022maxim}, we employ the Charbonnier loss $\mathcal{L}_{char}$~\cite{zamir2021multi}, the frequency loss $\mathcal{L}_{freq}$~\cite{jiang2021focal} and the edge loss $\mathcal{L}_{edge}$~\cite{jiang2020multi} to constrain scale-specific learning.
Furthermore, we also employ a $L_{1}$-norm to avoid color shift during the prediction of RGB by INR.
Based on one coarse and one fine feature grid, the total INR-related losses are calculated as follows:
\vspace{-2mm}
\begin{equation}
\vspace{-2mm}
\mathcal{L}_{inr}=\sum_{s=1}^2 \left\|\mathbf{I}_{s}-\mathbf{T}_{s}\right\|_1,
\end{equation}
where $\mathbf{I}_{s}$ and $\mathbf{T}_{s}$ denote $s$-scale reconstructed image of INR and $s$-scale target ground-truth image. The proposed loss function $\mathcal{L}_{total}$ for network training is defined as:
\vspace{-2mm}
\begin{equation}
\vspace{-2mm}
\mathcal{L}_{total}=\mathcal{L}_{char}+ \alpha_1 \mathcal{L}_{freq}+\alpha_2 \mathcal{L}_{edge}+\alpha_3 \mathcal{L}_{inr},
\end{equation}
where the scalar weights $\alpha_1$, $\alpha_2$ and $\alpha_3$ are empirically set to 0.01, 0.05 and 0.1, respectively.

\section{Experimental Results}
\label{sec:experiments}
\vspace{-2mm}
We first discuss the experimental settings of our proposed NeRD-Rain. Then we evaluate the effectiveness of our approach on both synthetic and real-world datasets.
More results are included in the supplemental material.
\vspace{-1mm}
\begin{figure*}[t]
	\footnotesize
	\centering
	\begin{tabular}{cccccc}
         \hspace{-2.5mm}
		\includegraphics[width=0.162\textwidth]{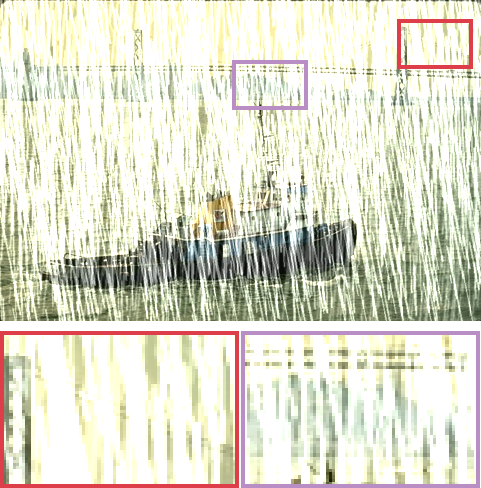}&\hspace{-4.1mm}
		\includegraphics[width=0.162\textwidth]{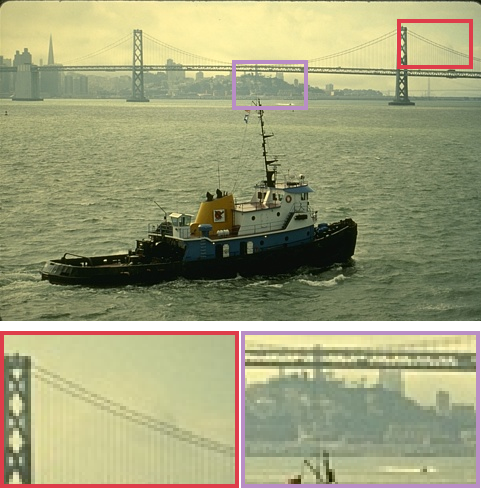}&\hspace{-4.1mm}
		\includegraphics[width=0.162\textwidth]{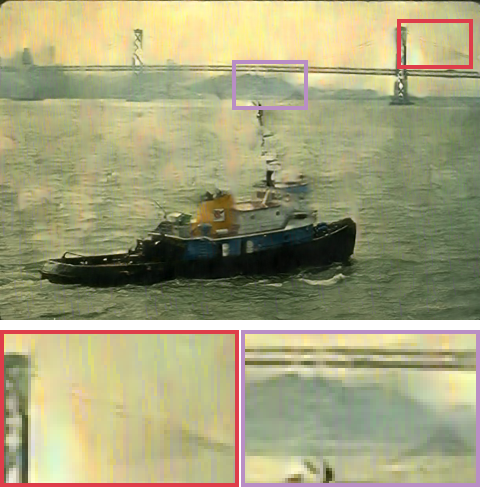}&\hspace{-4.1mm}
		\includegraphics[width=0.162\textwidth]{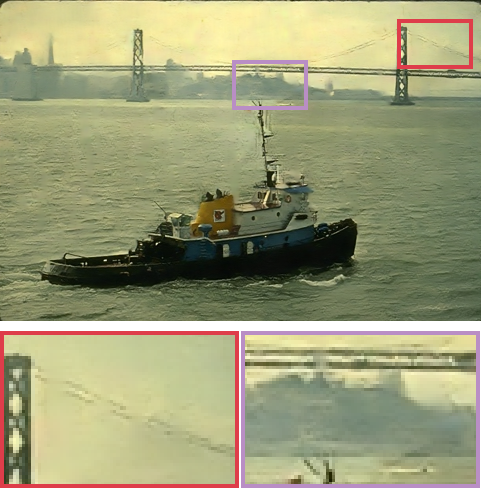}&\hspace{-4.1mm}
		\includegraphics[width=0.162\textwidth]{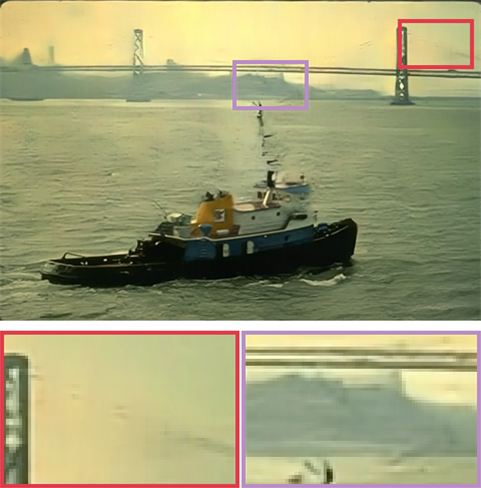}&\hspace{-4.1mm}
		\includegraphics[width=0.162\textwidth]{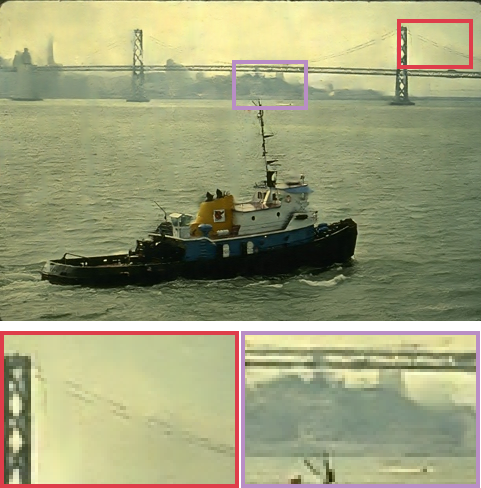}\\
		(a) Rainy input &\hspace{-4.1mm} (b) Ground truth  &\hspace{-4.1mm} (c) MSPFN \cite{jiang2020multi} &\hspace{-4.1mm} (d) RCDNet \cite{wang2020model} &\hspace{-4.1mm} (e) MPRNet \cite{zamir2021multi} &\hspace{-4.1mm} (f) DualGCN \cite{fu2021rain} \\
        \hspace{-2.5mm}
		\includegraphics[width=0.162\textwidth]{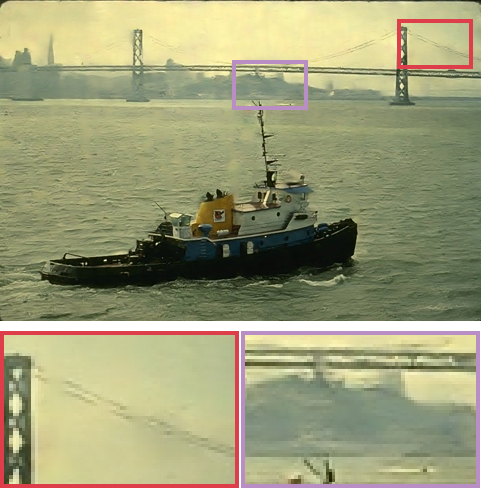}&\hspace{-4.1mm}
		\includegraphics[width=0.162\textwidth]{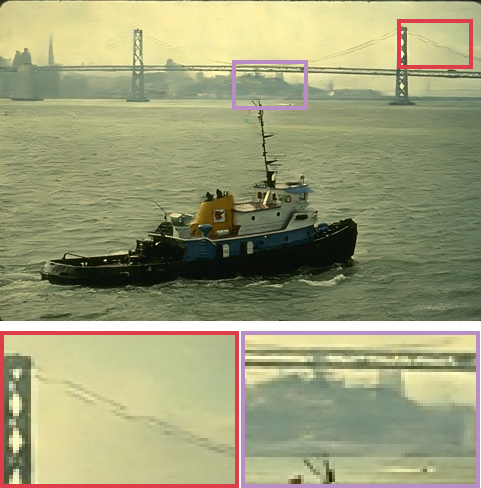}&\hspace{-4.1mm}
		\includegraphics[width=0.162\textwidth]{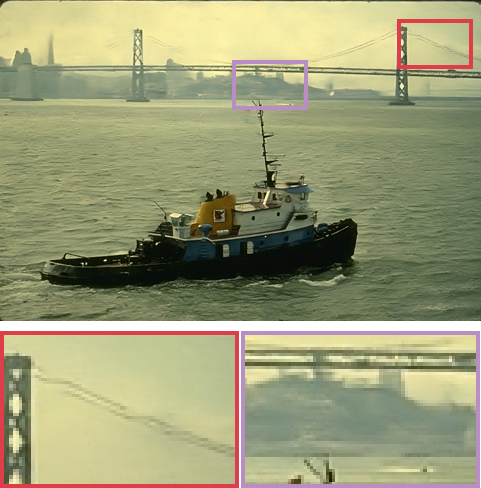}&\hspace{-4.1mm}
		\includegraphics[width=0.162\textwidth]{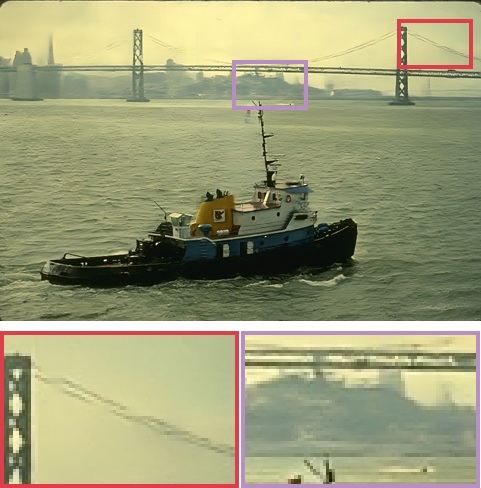}&\hspace{-4.1mm}
		\includegraphics[width=0.162\textwidth]{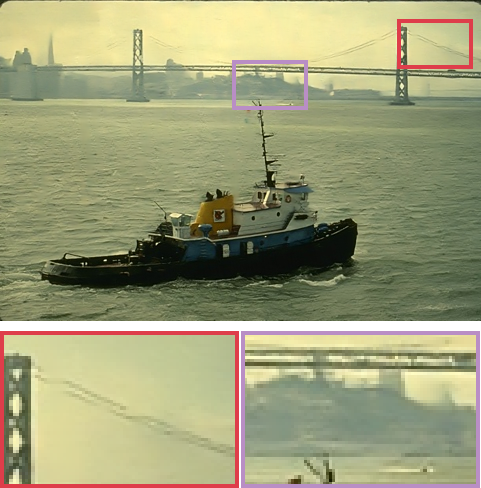}&\hspace{-4.1mm}
		\includegraphics[width=0.162\textwidth]{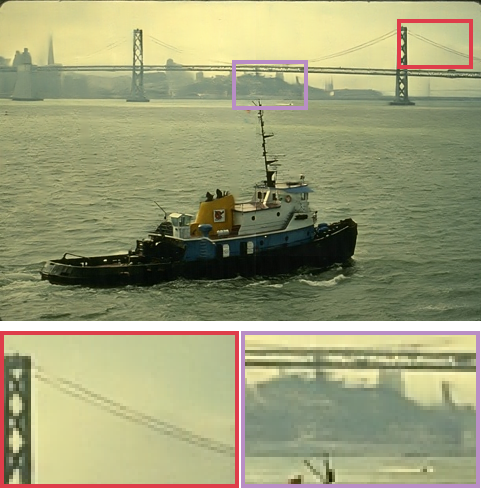}\\
		(g) SPDNet \cite{yi2021structure} &\hspace{-4.1mm} (h) Uformer \cite{wang2022uformer} &\hspace{-4.1mm} (i) Restormer \cite{zamir2022restormer} &\hspace{-4.1mm} (j) IDT  \cite{xiao2022image} &\hspace{-4.1mm} (k) DRSformer \cite{chen2023learning} &\hspace{-4.1mm} (l) Ours \\
	\end{tabular}
	\vspace{-3mm}
	\caption{Derained results on the Rain200H dataset~\cite{yang2017deep}. Compared with the derained results in (c)-(k), our method recovers a high-quality image with clearer details. Zooming in the figures offers a better view at the deraining capability.}
	\label{fig3}
	\vspace{-3mm}
\end{figure*}

\begin{figure*}[t]
	\footnotesize
	\centering
	\begin{tabular}{ccccccc}
        \hspace{-2.5mm}
		\includegraphics[width=0.138\textwidth]{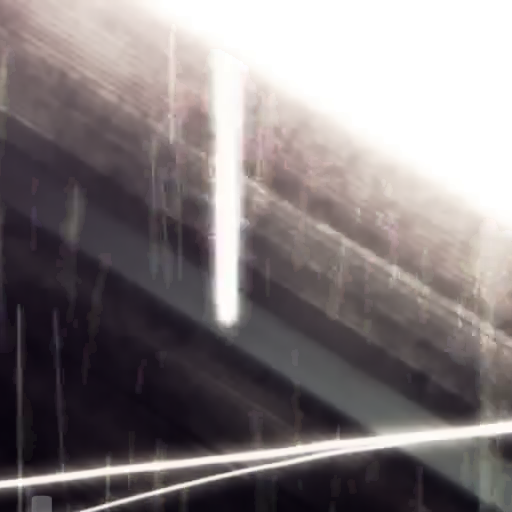}&\hspace{-4.1mm}
		\includegraphics[width=0.138\textwidth]{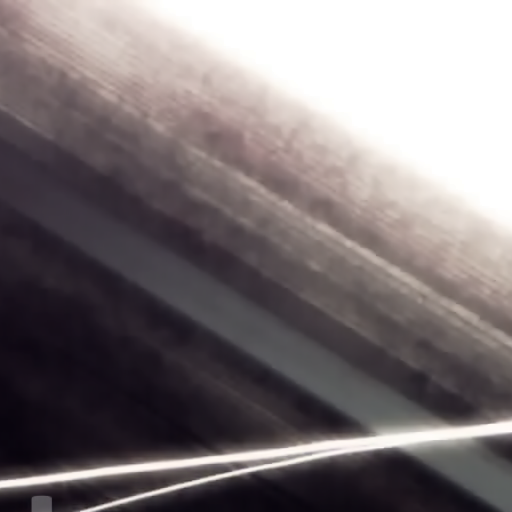}&\hspace{-4.1mm}
		\includegraphics[width=0.138\textwidth]{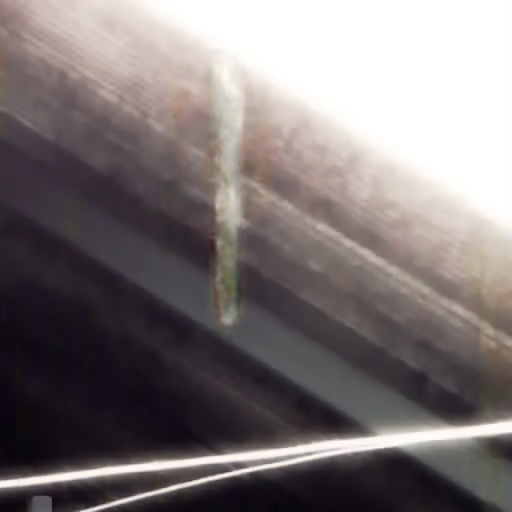}&\hspace{-4.1mm}
		\includegraphics[width=0.138\textwidth]{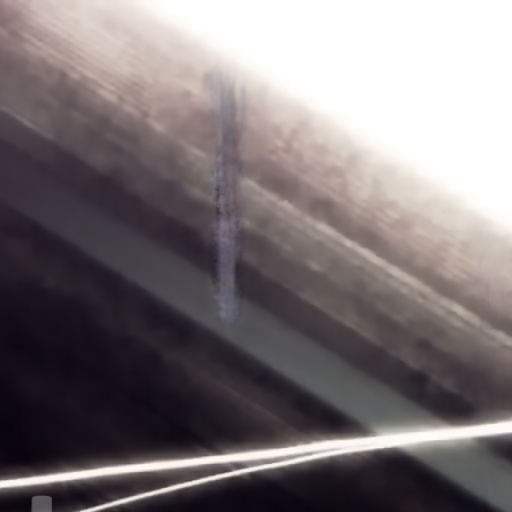}&\hspace{-4.1mm}
		\includegraphics[width=0.138\textwidth]{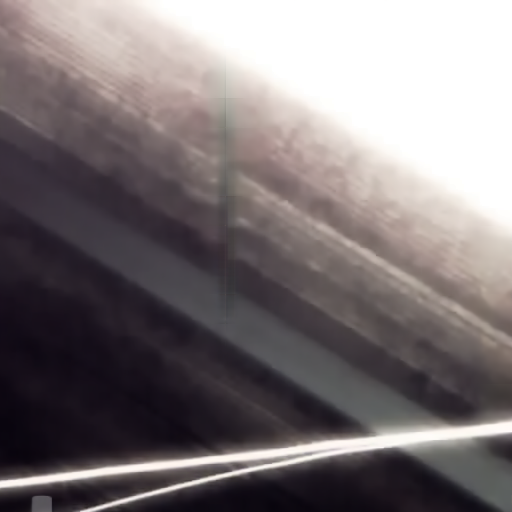}&\hspace{-4.1mm}
		\includegraphics[width=0.138\textwidth]{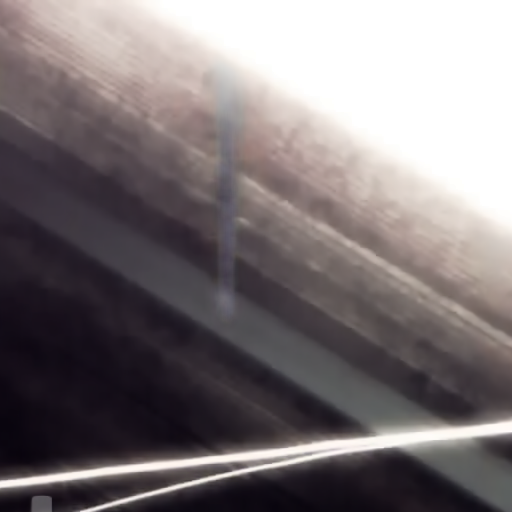}&\hspace{-4.1mm}
		\includegraphics[width=0.138\textwidth]{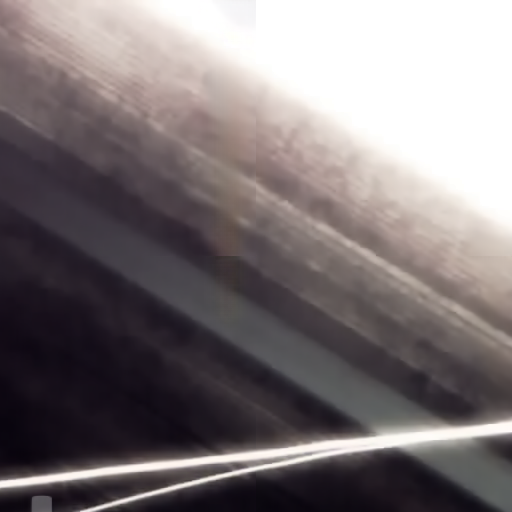} \\
        \hspace{-2.5mm}
		\includegraphics[width=0.138\textwidth]{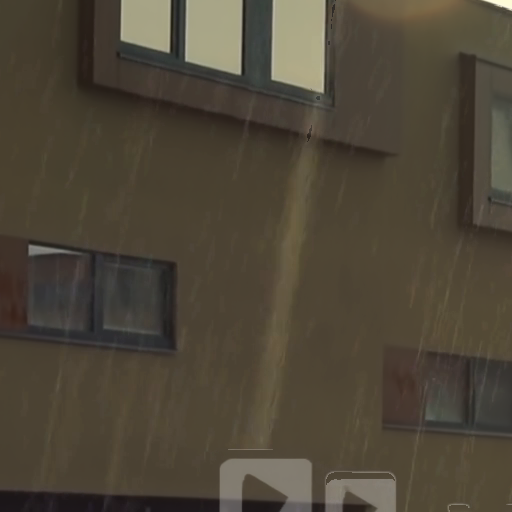}&\hspace{-4.1mm}
		\includegraphics[width=0.138\textwidth]{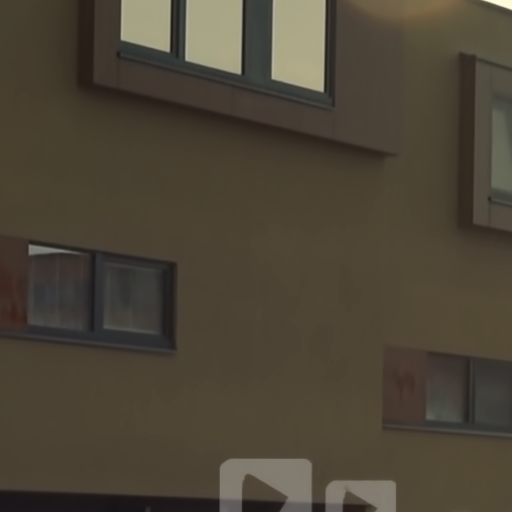}&\hspace{-4.1mm}
		\includegraphics[width=0.138\textwidth]{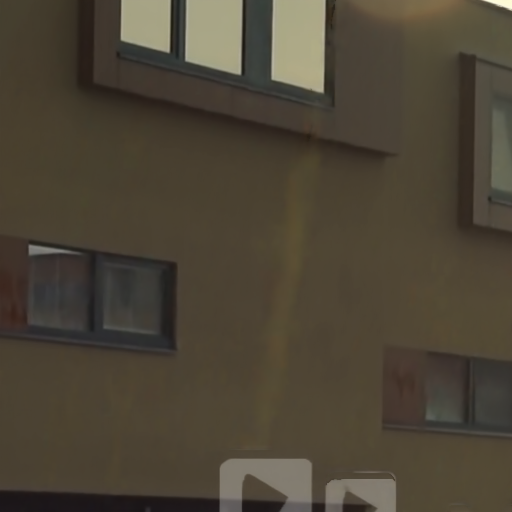}&\hspace{-4.1mm}
		\includegraphics[width=0.138\textwidth]{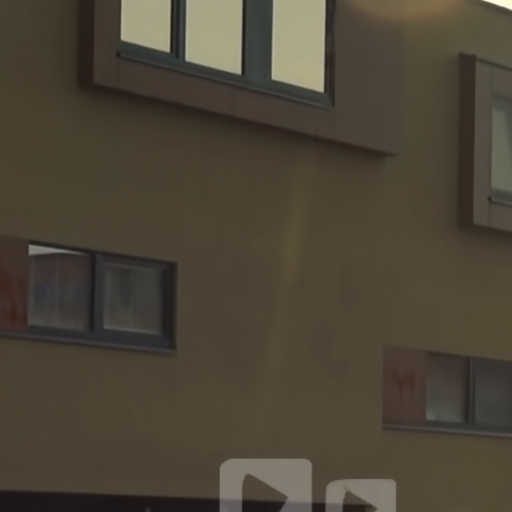}&\hspace{-4.1mm}
		\includegraphics[width=0.138\textwidth]{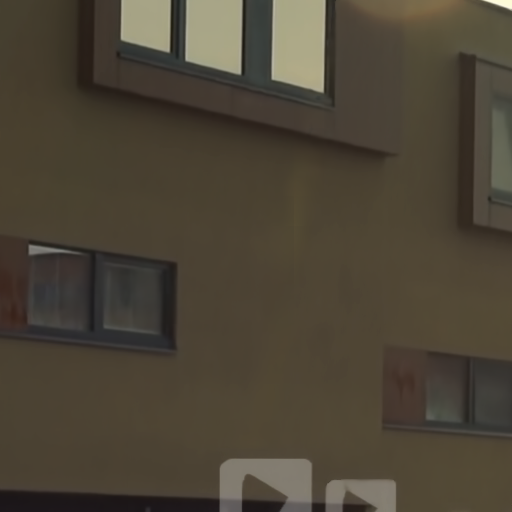}&\hspace{-4.1mm}
		\includegraphics[width=0.138\textwidth]{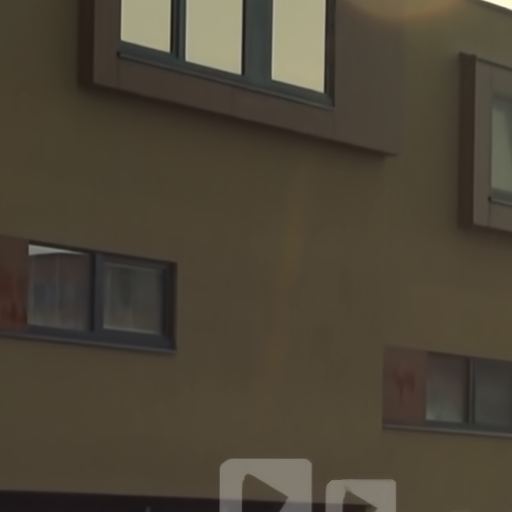}&\hspace{-4.1mm}
		\includegraphics[width=0.138\textwidth]{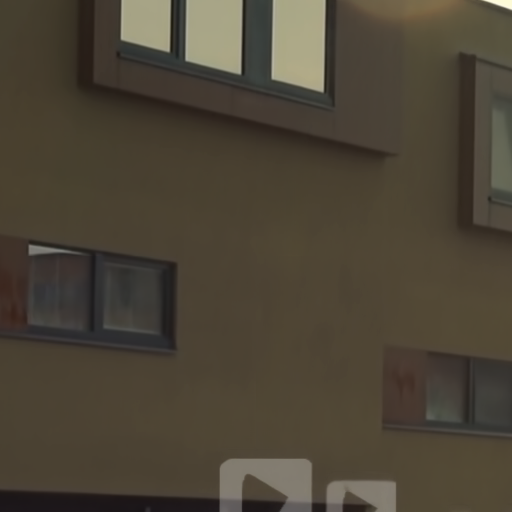} \\
		(a) Rainy input &\hspace{-4.1mm} (b) Ground truth  &\hspace{-4.1mm} (c) SPDNet \cite{yi2021structure} &\hspace{-4.1mm} (d) Restormer \cite{zamir2022restormer} &\hspace{-4.1mm} (e) IDT \cite{xiao2022image} &\hspace{-4.1mm} (f) DRSformer \cite{chen2023learning} &\hspace{-4.1mm} (g) Ours\\
	\end{tabular}
	\vspace{-3mm}
	\caption{Derained results on the SPA-Data \cite{wang2019spatial} dataset. Compared with the derained results in (c)-(f), our method recovers clearer images.}
	\label{fig4}
	\vspace{-4mm}
\end{figure*}

\subsection{Experimental settings}
\vspace{-1mm}
{\flushleft\textbf{Datasets and metrics}.}
We evaluate our approach on four commonly used synthetic benchmarks (\ie, Rain200L~\cite{yang2017deep}, Rain200H~\cite{yang2017deep}, DID-Data~\cite{zhang2018density}, and DDN-Data~\cite{fu2017removing}), and two recent real-world datasets (\ie, SPA-Data~\cite{wang2019spatial} and RE-RAIN~\cite{chen2023towards}). We follow the protocols of these benchmarks for training and testing.
%
To evaluate the quality of each derained image, we use PSNR \cite{PSNR} and SSIM \cite{wang2004image} as the evaluation metrics when the ground truth images are available, and calculate them based on the Y channel of the YCbCr color space, following the previous works~\cite{chen2023learning,chen2023towards}.

\vspace{-2mm}

{\flushleft\textbf{Implementation details}.}
In our proposed NeRD-Rain, each Transformer-based UNet adopts a 3-level encoder-decoder architecture.
From level-1 to level-3, the number of Transformer blocks is set to $[2,3,3]$, the number of self-attention heads in \cite{zamir2022restormer} is set to $[1,2,4]$, and the number of channels is set to $[48,96,192]$.
We introduce another variant, NeRD-Rain-S, by modifying the number of feature channels to $[32,64,128]$.
We implement our method based on the PyTorch framework and train it from scratch using a machine with one NVIDIA GeForce RTX 3090 GPU.
During the training, we use the Adam optimizer~\cite{kingma2014adam}.
%
The patch size is set to be $256 \times 256$ pixels and the batch size is set to be $1$.
We train the SPA-Data dataset for 10 epochs, the DID-Data and DDN-Data datasets for 200 epochs, and the Rain200L and Rain200H datasets for 600 epochs.
The same data augmentation method~\cite{mao2023intriguing} is adopted.
For training on the Rain200H dataset, the learning rate is initiallized as $2 \times 10^{-4}$, while for other benchmarks, it is initialized as $1 \times 10^{-4}$.
The final learning rate is gradually decreased to $1 \times 10^{-6}$ using a cosine annealing scheme~\cite{loshchilov2016sgdr}.

\vspace{-1mm}

\subsection{Comparisons with the state of the art}
\vspace{-2mm}
We compare our method with prior-based methods (DSC \cite{luo2015removing} and GMM \cite{li2016rain}), CNN-based approaches (DDN \cite{fu2017removing}, RESCAN \cite{li2018recurrent}, PReNet \cite{ren2019progressive}, MSPFN \cite{jiang2020multi}, RCDNet \cite{wang2020model}, MPRNet \cite{zamir2021multi}, DualGCN \cite{fu2021rain} and SPDNet \cite{yi2021structure}), and recent Transformer-based methods (Uformer \cite{wang2022uformer}, Restormer \cite{zamir2022restormer}, IDT \cite{xiao2022image}, and DRSformer \cite{chen2023learning}).

\vspace{-2mm}

{\flushleft\textbf{Evaluations on synthetic datasets}.}
Table \ref{table1} summarizes the quantitative evaluation results on the above-mentioned synthetic datasets \cite{yang2017deep,zhang2018density,fu2017removing}, where the derained images by our method have higher PSNR and SSIM values. For example, the PSNR values of our approach is at least 0.48 dB higher than DRSformer \cite{chen2023learning} on the Rain200L benchmark.

We further show some visual results on the Rain200H dataset in Figure \ref{fig3}.
The CNN-based methods \cite{jiang2020multi,wang2020model,zamir2021multi,fu2021rain,yi2021structure} do not recover clear images in heavy rainy scenarios.
The Transformer-based methods \cite{wang2022uformer,zamir2022restormer,xiao2022image,chen2023learning} are able to model the global contexts for image deraining.
However, some main structures, \eg, slender cable, are not recovered well.
Compared to existing Transformer-based methods that depend on single-scale rain appearance, our developed multi-scale Transformer is able to explore multi-scale representations of rain streaks, and generates clearer images with fine details and structures.

\vspace{-2mm}

\begin{figure*}[t]
	\footnotesize
	\centering
	\begin{tabular}{ccccccc}
 \hspace{-2.5mm}
		\includegraphics[width=0.139\textwidth]{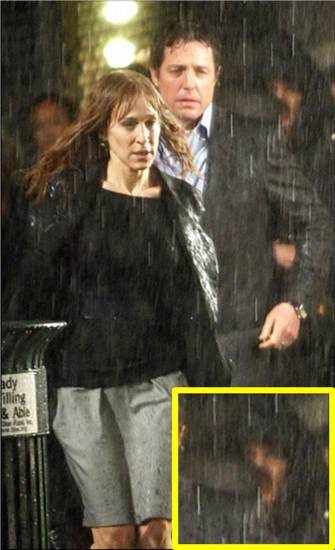}&\hspace{-4.1mm}
		\includegraphics[width=0.139\textwidth]{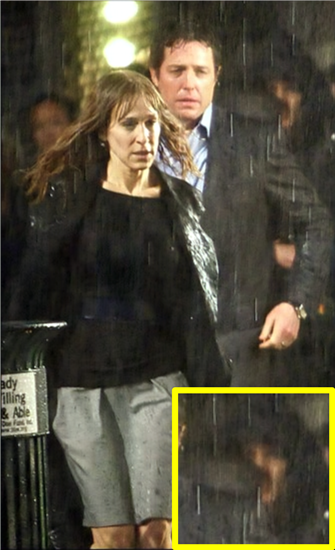}&\hspace{-4.1mm}
		\includegraphics[width=0.139\textwidth]{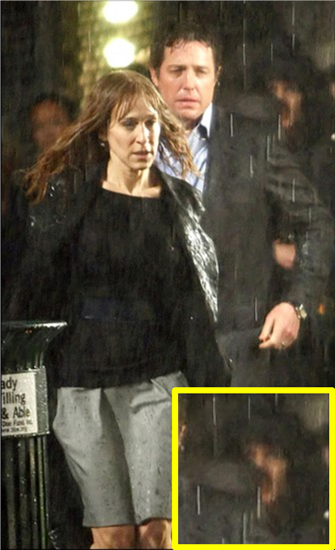}&\hspace{-4.1mm}
		\includegraphics[width=0.139\textwidth]{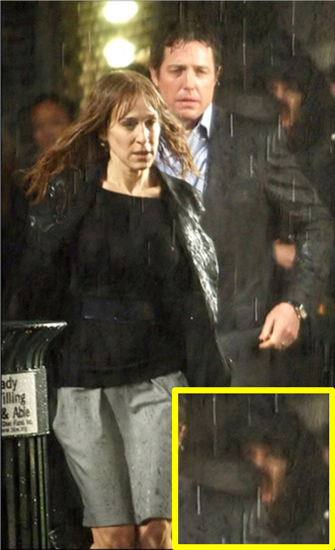}&\hspace{-4.1mm}
		\includegraphics[width=0.139\textwidth]{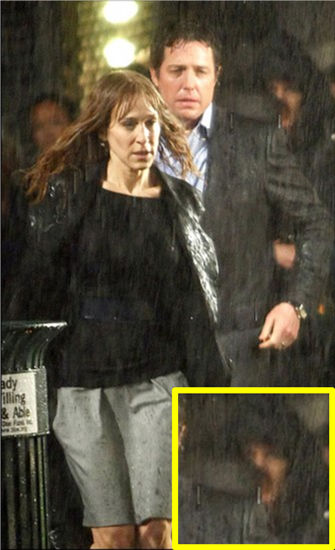}&\hspace{-4.1mm}
		\includegraphics[width=0.139\textwidth]{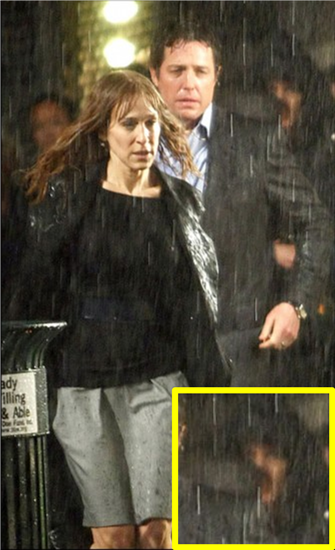}&\hspace{-4.1mm}
		\includegraphics[width=0.139\textwidth]{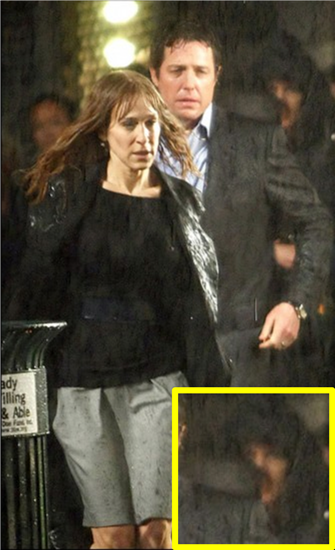}\\
		(a) Rainy input &\hspace{-4.1mm} (b) SPDNet \cite{yi2021structure}  &\hspace{-4.1mm} (c) Uformer \cite{wang2022uformer} &\hspace{-4.1mm} (d) Restormer \cite{zamir2022restormer} &\hspace{-4.1mm} (e) IDT \cite{xiao2022image} &\hspace{-4.1mm} (f) DRSformer \cite{chen2023learning} &\hspace{-4.1mm} (g) Ours
	\end{tabular}
	\vspace{-3mm}
	\caption{Derained results on a real-world rainy image from \cite{chen2023towards}. Compared with the derained results in (b)-(f), our method removes most rain streaks and recovers a clearer image. Zooming in the figures offers a better view at the deraining capability.}
	\label{fig5}
	\vspace{-3mm}
\end{figure*}

{\flushleft\textbf{Evaluations on real-world datasets}.}
We further evaluate our method on the challenging SPA-Data dataset~\cite{wang2019spatial}.
%
As provided in the last column of Table~\ref{table1}, our method outperforms the DRSformer method \cite{chen2023learning} by 1.04 dB on the SPA-Data dataset.
This indicates that our method can effectively deal with diverse types of spatially-varying real rain streaks.

Figure \ref{fig4} shows some visual comparisons of the evaluated methods, where our method generates better derained images.
In contrast, the recovery results of other methods still contain some undesired rain streaks residual.

We also evaluate our method using real captured rainy images from the RE-RAIN dataset~\cite{chen2023towards}, where ground truths are not available.
Figure~\ref{fig5} shows that most deep models are sensitive to spatially-long rain streaks and leave behind residual rain effect.
On the contrary, our method effectively removes random rain streaks and achieves better recovery results, indicating its ability to generalize well on real data.

\begin{table}[t]
	\centering
	\caption{Comparisons of model complexity against state-of-the-art methods. The size of the test image is $256 \times 256$ pixels. ``\#FLOPs'' and ``\#Params'' represent FLOPs (in G) and the number of trainable parameters (in M), respectively.}
	\vspace{-3mm}
	\resizebox{1.0\columnwidth}{!}{
		\begin{tabular}{c|cccc}
			\toprule
			Methods      & MSPFN \cite{jiang2020multi} & IPT \cite{chen2021pre}   & Uformer \cite{wang2022uformer} & Restormer \cite{zamir2022restormer} \\ \hline
			\#FLOPs (G)  & 595.5 & 1188     & 45.9    & 174.7     \\
			\#Params (M) & 13.35  & 115.5      & 50.88   & 26.12     \\ \hline
			Methods      & IDT \cite{xiao2022image}  & DRSformer  \cite{chen2023learning}& Ours-S   & Ours      \\ \hline
			\#FLOPs (G)  & 61.9  & 242.9     & 79.2    & 156.3     \\
			\#Params (M) & 16.41 & 33.65     & 10.53    & 22.89      \\ \bottomrule
		\end{tabular}
	}	
	\label{table2}
	\vspace{-4mm}
\end{table}
\vspace{-2mm}
{\flushleft\textbf{Model complexity}.}
We evaluate the model complexity of our method and state-of-the-art ones in terms of FLOPs and model parameters.
Table~\ref{table2} shows that our model, NeRD-Rain-S, has lower FLOPs value and fewer parameters while achieving competitive performance as shown in Table~\ref{table1}.

\vspace{-2mm}

\begin{table*}[t]
	\centering
	\caption{Ablation analysis on different variants of INR in our method, including four aspects: position, feature-grid, operation, and feature encoder. Here, all ablation models adopt the backbone of bidirectional multi-scale Transformer (BMT).}
	\vspace{-2mm}
	\resizebox{1.0\textwidth}{!}{
	\begin{tabular}{cccccccccccc}
	\toprule
	\multirow{3}{*}{Methods} & \multicolumn{8}{c}{Implicit Neural Representation (INR)}                                                                                                                 & \multirow{2}{*}{Backbone} & \multicolumn{2}{c}{\multirow{2}{*}{Metrics}} \\ \cline{2-9}
	& \multicolumn{2}{c}{Position} & \multicolumn{2}{c}{Feature grid} & \multicolumn{2}{c}{Operation}          & \multicolumn{2}{c}{Feature encoder}       &                           & \multicolumn{2}{c}{}                         \\ \cline{2-12}
	& Within branch   & Adjacent branch  & Fixed-scale     & Multi-scale    & Position encoding & Interpolation & Shared encoder & Separate encoder & BMT                       & PSNR                 & SSIM                  \\ \hline
	(a)                        & \XSolidBrush             & \XSolidBrush            & \XSolidBrush               & \XSolidBrush              & \XSolidBrush                 & \XSolidBrush             & \XSolidBrush              & \XSolidBrush                & \CheckmarkBold                         & 41.40                & 0.9896                \\
	(b)                        & \CheckmarkBold             & \XSolidBrush            & \XSolidBrush               & \CheckmarkBold              & \CheckmarkBold                 & \CheckmarkBold             & \XSolidBrush              & \CheckmarkBold                & \CheckmarkBold                         & 41.51                & 0.9897                \\
	(c)                        & \XSolidBrush             & \CheckmarkBold            & \XSolidBrush               & \CheckmarkBold              & \CheckmarkBold                 & \CheckmarkBold             & \CheckmarkBold              & \XSolidBrush                & \CheckmarkBold                         & \textbf{41.71}                & \textbf{0.9903}                \\
	(d)                        & \XSolidBrush             & \CheckmarkBold            & \CheckmarkBold               & \XSolidBrush              & \CheckmarkBold                 & \CheckmarkBold             & \CheckmarkBold              & \XSolidBrush                & \CheckmarkBold                         & 41.47                & 0.9897                \\
	(e)                        & \XSolidBrush             & \CheckmarkBold            & \XSolidBrush               & \CheckmarkBold              & \XSolidBrush                 & \CheckmarkBold             & \CheckmarkBold              & \XSolidBrush                & \CheckmarkBold                         & 41.50                & 0.9899                \\
	(f)                        & \XSolidBrush             & \CheckmarkBold            & \XSolidBrush               & \CheckmarkBold              & \CheckmarkBold                 & \XSolidBrush             & \CheckmarkBold              & \XSolidBrush                & \CheckmarkBold                         & 41.58                &  0.9900               \\
	(g)                        & \XSolidBrush             & \CheckmarkBold            & \XSolidBrush               & \CheckmarkBold              & \CheckmarkBold                 & \CheckmarkBold             & \XSolidBrush              & \CheckmarkBold                & \CheckmarkBold                         & 41.62                & 0.9901                \\ \bottomrule
\end{tabular}
	}
	\vspace{-4mm}
	\label{table3}	
\end{table*}

\section{Analysis and Discussion}
\label{sec:analysis}
\vspace{-2mm}
To better understand how the proposed approach solves image deraining, we examine the effect of the main component by conducting ablation studies. We train our method and all the alternative baselines using the same settings for fairness.

\begin{table}[t]\scriptsize
	\centering
	\caption{Ablation analysis on the multi-scale configuration using different numbers of UNets. Here, $S_{1}$, $S_{2}$, and $S_{3}$ represent $1/4$, $1/2$ and full image scale, respectively.}
	\vspace{-3mm}
	\resizebox{1.0\columnwidth}{!}{
	\begin{tabular}{cccccc}
	\toprule
	Methods & ~~$S_{1}$~~ & ~~$S_{2}$~~ & ~~$S_{3}$~~ & ~~PSNR~~  & ~~SSIM~~   \\ \hline
	M222      & 2  & 2  & 2  & 41.42 & 0.9896 \\
	M321      & 3  & 2  & 1  & 40.70 & 0.9882 \\
	M023      & 0  & 2  & 3  & 41.63 & 0.9901 \\
	M123 (Ours)     & 1  & 2  & 3  & \textbf{41.71} & \textbf{0.9903} \\ \bottomrule
    \end{tabular}
	}	
\label{table4}
\vspace{-4mm}
\end{table}

\vspace{-3mm}

{\flushleft\textbf{Effect of multi-scale configuration}.}
Our developed multi-scale Transformer focuses on incorporating finer scales with deeper architectures to better remove spatially-varying rain streaks.
To demonstrate the effectiveness of this formulation, we analyze the effect of different multi-scale configurations on the Rain200L dataset.
As shown in Table~\ref{table4}, these variants differ in the number of UNets at different scales while sharing the same network architecture for the UNets.
%
%
Compared to treating each scale equally (\ie, M222), our method (\ie, M123) can better reconstruct scale-specific feature and improve potential image restoration quality.

\vspace{-2mm}

{\flushleft\textbf{Effectiveness of INR branch}.}
To analyze the effectiveness of INR branch, we conduct ablation experiments based on different variants in Table \ref{table3}.
All variants are trained on the Rain200L dataset.
We compare with the baseline without using INR (\ie, model (a)). In contrast, our model (c) exhibits better performance, especially in real scenes where the capability for effective rain removal is more pronounced (Figure~\ref{fig6}(b) and (d), see the supplemental material for more examples).
The approach without INR is sensitive to rain streaks of different scales, while our method successfully removes diverse rain streaks.
This confirms that the learned representations from our INR better facilitates rain removal.

To understand the effect of such INR, we first visualize the output of INR from our network, and pixel value distribution of images.
In Figure~\ref{fig8}, we find that INR reduces the high intensity pixel values of rainy images and generates rain-free images.
As mentioned in Section~\ref{sec:3.1}, we attribute this natural ability to the fact that INR tends to bias towards learning low-frequency image contents \cite{lee2022local}, while rain occupies the high-frequency part of the image.

Then, we further analyze the effect of implicit interpolation and position encoding in INR.
Figure~\ref{fig7}(b) shows that INR fails to remove undesired high-frequency rain streaks without using implicit interpolation operations.
Meanwhile, the output of INR without using position encoding are prone to losing background details, as shown in Figure~\ref{fig7}(c).
All in all, the former aims to find the local latent embedding to reconstruct RGB values of the clear image, while the latter focuses on encoding coordinates with the frequency information to recover fine-grained background details.

\begin{figure}[!t]\footnotesize
	\centering
	\begin{tabular}{cccc}
		\hspace{-2mm}
		\includegraphics[width=0.24\linewidth]{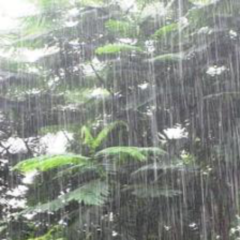} &\hspace{-4mm}
		\includegraphics[width=0.24\linewidth]{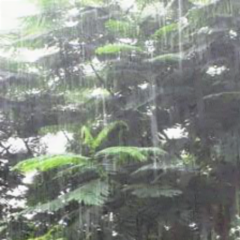} &\hspace{-4mm}
		\includegraphics[width=0.24\linewidth]{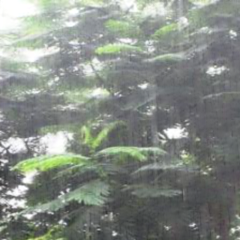} &\hspace{-4mm}
		\includegraphics[width=0.24\linewidth]{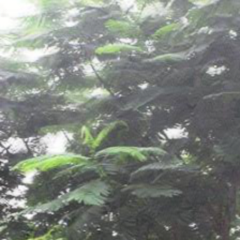}  \\
		(a) Rainy input &\hspace{-4mm}  (b) w/o INR & \hspace{-4mm} (c) w/o SE & \hspace{-4mm} (d) Ours
	\end{tabular}
	\vspace{-3mm}
	\caption{Ablation qualitative comparison on a real-world rainy image. ``SE'' denotes a shared encoder in our method.}
	\vspace{-6mm}
	\label{fig6}
\end{figure}

\begin{figure*}[t]
	\footnotesize
	\centering
	\begin{tabular}{cccccc}
		\hspace{-2mm}
		\includegraphics[width=0.162\textwidth]{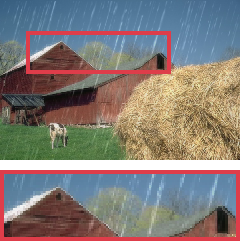}&\hspace{-4.1mm}
		\includegraphics[width=0.162\textwidth]{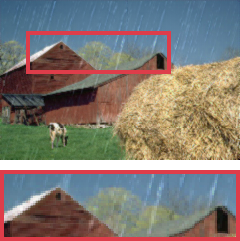}&\hspace{-4.1mm}
		\includegraphics[width=0.162\textwidth]{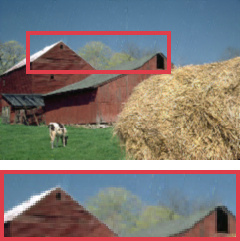}&\hspace{-4.1mm}
		\includegraphics[width=0.162\textwidth]{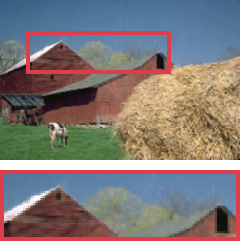}&\hspace{-4.1mm}
		\includegraphics[width=0.162\textwidth]{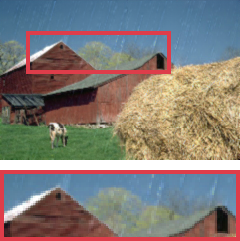}&\hspace{-4.1mm}
		\includegraphics[width=0.162\textwidth]{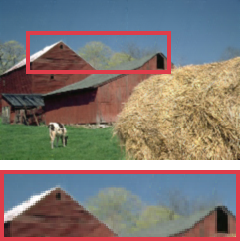}\\
		(a) Rainy input &\hspace{-4.1mm} (b) w/o Interpolation  &\hspace{-4.1mm} (c) w/o PE &\hspace{-4.1mm} (d) w/ Within branch &\hspace{-4.1mm} (e) w/ Fixed scale &\hspace{-4.1mm} (f) Ours \\
	\end{tabular}
	\vspace{-3mm}
	\caption{Visual quality comparison of output results of INR in different variants. Our INR branch generates a clearer image using all the designs we consider, which further provides better guidance for the input of the next scale. ``PE''denotes the position encoding operation.}
	\label{fig7}
	\vspace{-3mm}
\end{figure*}

\begin{figure}[!t]\footnotesize
	\centering
	\begin{tabular}{cc}
 \hspace{-2mm}
		\includegraphics[width=0.51\linewidth]{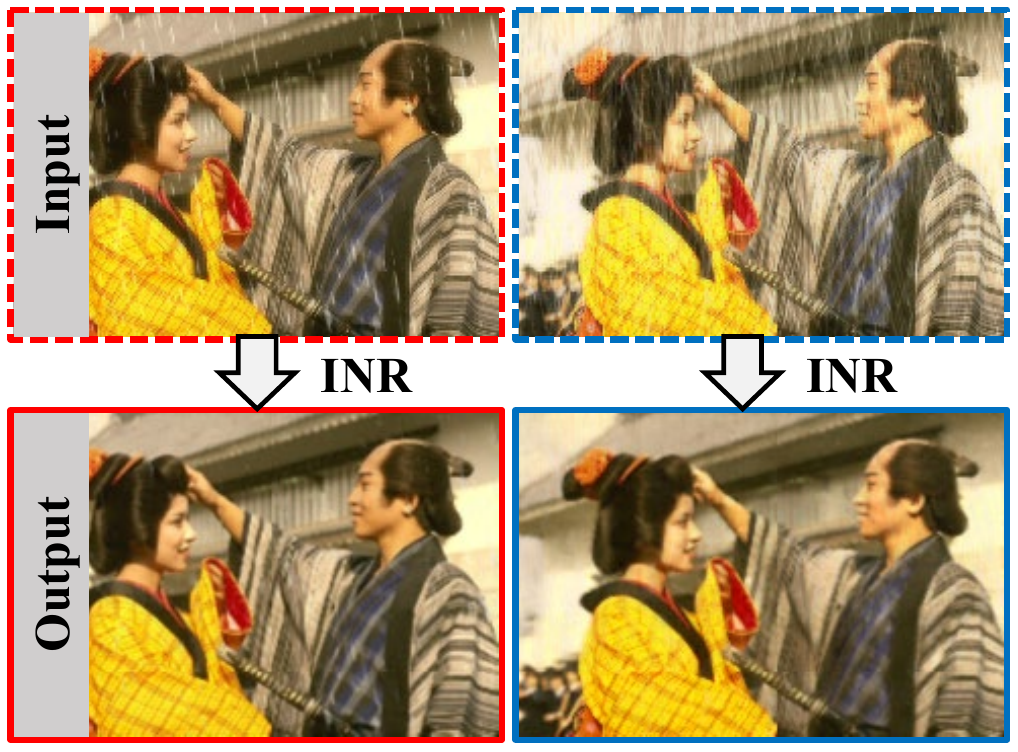} &\hspace{-3mm}
		\includegraphics[width=0.45\linewidth]{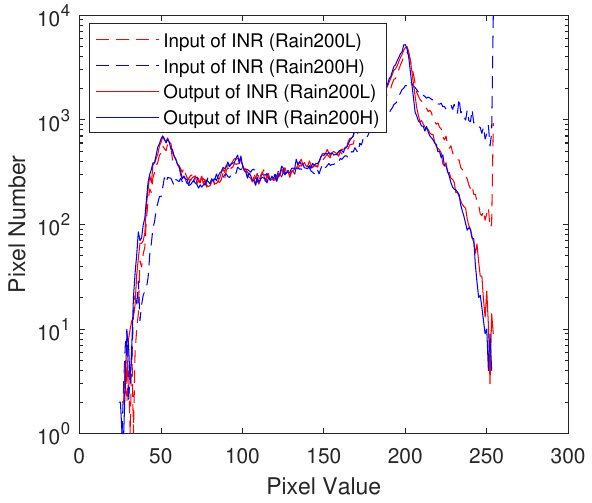}  \\
		(a) Visualized output results of INR &\hspace{-4mm}  (b) Pixel value distribution
	\end{tabular}
	\vspace{-3mm}
	\caption{Comparions between the input images and the results of INR. One can see that INR reduces the high intensity pixel values of white rain streaks and reconstructs the latent rain-free images.}
	\vspace{-5mm}
	\label{fig8}
\end{figure}

Next, we also evaluate the effect of the position of INR in our framework.
Compared to embedding the INR within the Transformer branch (\ie, model (b) in Table~\ref{table3}), our method inserts INR between adjacent scales to naturally shares information across scales, thereby improving quantitative and qualitative results (Figure~\ref{fig7}(d)).
Furthermore, we note that our method with multi-scale feature grid boosts the representation capacity of INR, which benefits from observing rain appearance at different resolutions (Figure \ref{fig7}(e) and (f)).

Finally, we demonstrate the effectiveness of shared encoder on image deraining.
Here, we construct separate encoders for each INR as a comparison.
For fair comparison, we set the same network architecture for these encoders as our method.
By comparing model (c) and (g) in Table~\ref{table3}, our method achieves higher quantitative results with fewer network parameters.
Figure \ref{fig6}(c) and (d) also demonstrate that using the shared encoder generates much clearer images.
In contrast, the method with the separate encoder still has residual rain streaks in the recovery results.
This further indicates that our formulation is robust to complex real-world scenarios.
The utilization of shared encoders forms a more compact closed-loop framework, enabling the learned degradation representations to better facilitate rain removal.

\vspace{-1mm}

\begin{figure}[!t]\footnotesize
	\centering
	\begin{tabular}{cccc}
		\includegraphics[width=0.235\linewidth]{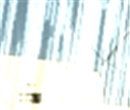} &\hspace{-4mm}
		\includegraphics[width=0.235\linewidth]{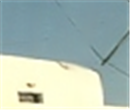} &\hspace{-4mm}
		\includegraphics[width=0.235\linewidth]{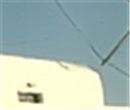} &\hspace{-4mm}
		\includegraphics[width=0.235\linewidth]{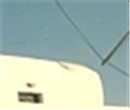}  \\
		(a) Rainy patch &\hspace{-4mm}  (b) w/o BFPU & \hspace{-4mm} (c)  w/ Concat & \hspace{-4mm} (d) Ours
	\end{tabular}
	\vspace{-2mm}
	\caption{Ablation qualitative comparison on the proposed BFPU.}
	\vspace{-4mm}
	\label{fig9}
\end{figure}

{\flushleft\textbf{Effectiveness of bidirectional branch}.}
The BFPU in the bidirectional branch is used to better explore bidirectional information in our multi-scale Transformer for better image restoration.
To demonstrate the effectiveness of this branch, we remove this component and investigate its influence in Table~\ref{table5}.
We note that the BFPU achieves a PSNR gain of 0.10 dB over unidirectional propagation (\ie, w/o BFPU) on the Rain200H dataset.
Compared to direct feature concatenation, our bidirectional branch can dynamically aggregate richer features to facilitate image restoration.
Figure~\ref{fig9} also shows that our method generates much clearer details.

\vspace{-1mm}

{\flushleft\textbf{Extension to various architectures.}.}
We extend our network architecture to CNN-based U-Net to demonstrate the scalability of the proposed framework. Here, we choose the U-Net module in MPRNet~\cite{zamir2021multi} as the baseline. Table~\ref{table6} reports the results trained on the Rain200L benchmark. Our method still achieves competitive performance, indicating the effectiveness of our method in various architectures.

\vspace{-1mm}

{\flushleft\textbf{Limitations and failure cases}.}
Although our NeRD-Rain achieves favorable performance on several image deraining benchmarks, its training time is relatively longer compared to other multi-scale architectures. This is mainly due to the time-consuming optimization process involved in INR.
Future work will apply the model pruning or early stopping scheme to improve training speed while maintaining performance~\cite{leclerc2023ffcv}.
Furthermore, our method fails to handle the veiling effect (\ie, mist) in complex rainy environments.

\vspace{-1mm}

\begin{table}[t]\footnotesize
	\centering
	\caption{Ablation quantitative comparison on the proposed BFPU.}
	\vspace{-3mm}
	\resizebox{1.0\columnwidth}{!}{
		\begin{tabular}{cccc}
			\toprule
			Methods & w/o BFPU & w/ Concat & w/ BFPU (Ours)  \\ \hline
			PSNR / SSIM   & 32.30 / 0.9346   & 32.31 / 0.9360   & \textbf{32.40} / \textbf{0.9373} \\ \bottomrule
		\end{tabular}
	}
	\vspace{-2mm}
	\label{table5}	
\end{table}

\begin{table}[t]\footnotesize
	\centering
	\caption{Extension to CNN-based U-Net~\cite{zamir2021multi} in our NeRD-Rain.}
	\vspace{-3mm}
	\resizebox{1.0\columnwidth}{!}{
		\begin{tabular}{cccc}
			\toprule
			Methods & MPRNet~\cite{zamir2021multi} & DRSformer~\cite{chen2023learning} & Ours  \\ \hline
			PSNR / SSIM   & 39.47 / 0.9825   & 41.23 / \textbf{0.9894}   & \textbf{41.34} / 0.9893 \\ \bottomrule
		\end{tabular}
	}
	\vspace{-2mm}
	\label{table6}	
\end{table}

\section{Concluding Remarks}
\label{sec:conclusion}
\vspace{-2mm}
We have presented an effective multi-scale  Transformer network for single image deraining.
To better explore common rain degradation features, we incorporate coordinated-based implicit neural representation between adjacent scales, enabling the learned features can better facilitate rain removal and improve the robustness of the model in complex scenarios.
To enhance the interactions among features of different scales in a collaborative manner, we also introduce a simple yet effective bidirectional feedback propagation operation into our multi-scale Transformer by performing coarse-to-fine and fine-to-coarse information communication.
By formulating the proposed method into an end-to-end trainable model, we show that it performs favorably against the state-of-the-art methods on both synthetic and real benchmarks.

\vspace{-2mm}
{\flushleft\textbf{Acknowledgements}.} This work has been partly supported by the National Natural Science Foundation of China (Nos. U22B2049, 62272233, 62332010), and the Postgraduate Research \& Practice Innovation Program of Jiangsu Province (No. KYCX23\_0486).

\clearpage
{
    \small
    \bibliographystyle{ieeenat_fullname}
    \bibliography{main}
}

\end{document}